\documentclass{article}
\usepackage{ijcai20}

\usepackage{times}
\usepackage{soul}
\usepackage{url}

\usepackage{hyperref}
\hypersetup{hidelinks}
\usepackage[utf8]{inputenc}
\usepackage[small]{caption}
\usepackage{graphicx}
\usepackage{amsmath}
\usepackage{amsthm}
\usepackage{amssymb}
\usepackage{booktabs}
\usepackage{algorithm}
\usepackage{algorithmic}
\usepackage{multirow}
\usepackage{graphicx}
\usepackage{subcaption}
\usepackage{mwe}
\title{Federated Learning System without Model Sharing through Integration of Dimensional Reduced Data Representations}

\author{
Anna Bogdanova$^1$\footnote{Contact Author}\and
Akie Nakai$^3$\and
Yukihiko Okada$^{1,3}$ \and
Akira Imakura$^{1,2}$ and
Tetsuya Sakurai$^{1,2}$\\
\affiliations
$^1$Center for Artificial Intelligence Research, University of Tsukuba\\
$^2$Department of Computer Science, University of Tsukuba\\
$^3$Department of Policy and Planning Sciences, University of Tsukuba\\
\emails
bogdanova.anna.fw@u.tsukuba.ac.jp,
s1920562@s.tsukuba.ac.jp,
okayu@sk.tsukuba.ac.jp,
\{imakura, sakurai\}@cs.tsukuba.ac.jp,
}

\begin{document}

\maketitle


\begin{abstract}
Dimensionality Reduction is a commonly used element in a machine learning pipeline that helps to extract important features from high-dimensional data. In this work, we explore an alternative federated learning system that enables integration of dimensionality reduced representations of distributed data prior to a supervised learning task, thus avoiding model sharing among the parties. We compare the performance of this approach on image classification tasks to three alternative frameworks: centralized machine learning, individual machine learning, and Federated Averaging, and analyze potential use cases for a federated learning system without model sharing. Our results show that our approach can achieve similar accuracy as Federated Averaging and performs better than Federated Averaging in a small-user setting.
\end{abstract}

\section{Introduction}
\subsection{Motivation}
In the modern data-driven economy and governance, combining multiple data sources for training machine learning models can lead to ground-breaking discoveries, novel services, and mitigation of public safety risks. These prospects motivate industries and governments to invest in developing data integration technologies and policies. 


However, even well-motivated data integration policies are rising concerns about the possible misuse of personal data and violation of data privacy rights, as well as corporate secrecy. New legislative provisions, such as the EU's General Data Protection Regulation (2018), and California's Consumer Privacy Act (2018), introduced essential measures to protect the personal data of users while also creating barriers to potentially useful data flows between industries and across borders. This trade-off poses the next big challenge for the research community to provide solutions that enable collaborative machine learning while protecting user privacy. 

Although multiple viable solutions and workflows have emerged over the recent years, there are still practical barriers to their implementation, including the lack of incentives, lack of necessary infrastructure, or unfitness of available solutions to particular needs. One specific obstacle to collaborative machine learning deployment is the necessity to share the resulting model with collaborators.

There are several use cases where sharing one machine learning model is not desirable. 
\begin{enumerate}
	\item In a scenario of semi-honest or adversarial participants, the shared model can be susceptible to model inversion attacks or model poisoning attacks. 
	\item In a private sector collaboration, machine learning models can be products of in-house know-how and costly development, so the prospects of using one shared model can disincentivize potential collaborators. 
	\item Collaborative training of one shared model requires iterative communication between the participants, which can be undesirable for security reasons or the absence of necessary infrastructure.
\end{enumerate}

Hence, our motivation in this work is to expand the vocabulary of current solutions by exploring an alternative framework that enables collaborative machine learning without model sharing.

In particular, in this paper, we analyze practical applications of an alternative method of distributed data analysis without model sharing, called Data Collaboration, by testing its performance on common image classification tasks. We further compare it to the Federated Averaging algorithm, which is the most commonly used among emerging Federated Learning Systems. We believe that the Data Collaboration method presented here can open a discussion towards filling in this missing link in practical applications of various Federated Learning Systems.  

\subsection{Related Work}
While it is a common practice to ensure a certain level of privacy by perturbing original data or stripping it of personal identifiers, contemporary methods of data analysis and the explosion of data call for more advanced solutions, as multiple studies demonstrated the unsafety of data anonymization \cite{rocher2019estimating}, \cite{de2013unique}.

A more recent trend has been to develop methods and approaches to collaboratively training machine learning models over distributed data. The method that has gained the most traction recently is Federated Learning, introduced by researchers at Google in 2016 \cite{mcmahan2016communication}. It features the Federated Averaging algorithm, which allows iteratively updating the centralized Neural Network model with averaged model weights trained locally at users' devices. Some of the known advantages of Federated Averaging is communication efficiency and robustness to user participation, when training convergence can be achieved even with multiple users dropping out of the network. In addition to training data not being accessible to the central server, further privacy-preserving measures were introduced to prevent privacy leakage from gradient updates. Thus, Secure Aggregation technique \cite{bonawitz2017practical} ensures that the server can never access individual model updates, and Differentially Private Federated Learning protects users from Membership Inference Attacks on the trained model \cite{geyer2017differentially}. Still, some open privacy and security issues remain to be addressed by the research community \cite{bagdasaryan2018backdoor}. 

The diversity of emerging variations and application of the Federated Learning approach became known as Federated Learning Systems \cite{li2019federated}, and they encompass different privacy mechanisms \cite{ryffel2018generic}, \cite{riazi2018chameleon}, various methods of data partitioning \cite{liu2018secure}, as well as specialized distributed learning algorithms \cite{vaidya2003privacy}, \cite{hamm2016learning}. One characteristic feature of all federated learning methods is that participants obtain a shared machine learning model, which they later use to process new data locally. 

In this paper, we present an alternative federated learning system without model sharing, previously introduced in \cite{imakura2020data}, \cite{ye2019distributed}, and \cite{imakura2020}, and analyze practical applications of this approach by testing its performance on common image classification tasks. 
This method relies on sharing and integrating intermediate representations of private data locally produced by each user by a secret dimensionality reduction function. Without knowledge of this function, private data cannot be restored or approximated. Although the method currently does not offer formal privacy guarantees, privacy-preserving properties of dimensionality reduction were previously explored elsewhere \cite{tai2018exploring}, \cite{nguyen2020autogan}. Moreover, formal privacy guarantees were demonstrated for particular methods of dimensionality reduction, such as non-metric multidimensional scaling (MDS) \cite{alotaibi2012non}, differential-private Principal Component Analysis, and differential-private Linear Discriminant Analysis \cite{jiang2013differential}, which could be applied within Data Collaboration framework for enhanced privacy protection.

\subsection{Paper Overview}
The following chapter gives a detailed description of the Data Collaboration method. Next, in Chapter 3, we present experiments with varying Data Collaboration parameters and comparing the results of image classification tasks with centralized learning, individual learning and conventional federated learning. Then, in the Discussion section, we analyse the experimental results and present several use-cases for the practical application of Data Collaboration method.

\section{Method description}
Data Collaboration Analysis, proposed in \cite{imakura2020data} and \cite{imakura2020}, is a method of computing a unified low-dimensional representation of distributed data, which can further be used for various supervised and unsupervised machine learning tasks independently by each party. For this paper, we consider supervised machine learning for image classification tasks. 
The proposed Data Collaboration setting consists of two or more participating institutions, each having their private datasets with target labels, and a semi-trusted server that computes unified data representation. In addition, participants jointly produce random data values called 'anchor data'. Each participant then computes a dimensionality reduction function of their choice and shares the transformed private data and anchor data with the server. Since the transformation functions remain private, original data cannot be approximated from the projections obtained by the server. The server uses anchor data to compute individual mapping functions that project intermediate representations into one space suitable for joint analysis.

More formally, let $d$ be the number of users, $X_i \in {R}^{n_i{\times}m}$  be the data sample of each user $i$, where $n_i$ is the sample size and $m$ is the number of features. We also let $L_i \in {R}^{n_i{\times}l}$, and $Y_i \in {R}^{s_i{\times}m}$ be the target labels and test dataset in the $i$th institution.
Each institution constructs lower dimensional representation of their data,%
\begin{align}
    \tilde{X}_i = f_i(X_i) \in {R}^{n_i{\times}l_i}\quad (0<l_i\leq{m}),
\end{align}%
where $f_i$ is any linear or non linear dimensionality reduction function uniquely parametrized by the local dataset, and the target dimension $l_i$ may vary among the users. Because of that, even if the parties share the same data sample $x$, its mapped representation will differ in each institution:%

\begin{align}
    f_i(x)\ne{f_j(x)}\quad (i\ne{j}).
\end{align}%
Therefore, in order to analyze centralized intermediate representations (IR) as one dataset we must find transformations $g_i$ such that%
\begin{align}
    g_i(f_i(x)){\approx}g_j(f_j(x))\quad (i\ne{j}).
\end{align}

Then, applying transformation function $g_i$ to each intermediate representation,%
\begin{align}
    \hat{X}_i = g_i(\tilde{X}_i) \in \mathbb{R}^{n_i{\times}l},
\end{align}
we obtain $\hat{X} = [\hat{X}_1^T, \hat{X}_2^T,..., \hat{X}_d^T]^T\in{R}^{n{\times}l}$, which has the property of preserving relationship among the samples of the original data, and can be analyzed as one dataset.

In order to construct the unifying transformation functions $g_i$, a sharable data, called \textit{anchors} is introduced, $X^{anc}\in\mathbb{R}^{r{\times}m}$, which can be either a randomly constructed dummy data, augmented or public data. Anchors are transformed by $f_i$ at each institution and their intermediate representations $\tilde{X}^{anc}_i\in\mathbb{R}^{r{\times}l_i}$ are shared with the server. 
Next, we set the target for the collaborative representation of the anchor data $Z\in{R}^{r{\times}l}$, such that%
\begin{align}
    \min_{G_1,G_2,...,G_d,Z}\sum_{i=1}^{d}||Z-g_i(\tilde{X}_i^{anc})||_F^2.
\end{align}
For the case that $g_i$ is linear, i.e., $\hat{X}_i = g_i(\tilde{X}_i) = \tilde{X}_iG_i$ with $G_i \in\mathbb{R}^{l_i{\times}l}$,
this minimization problem can be solved by singular value decomposition (SVD) of the matrix of $\tilde{X}^{anc}:=[\tilde{X}^{anc}_1, \tilde{X}^{anc}_2,...,\tilde{X}^{anc}_d] $,%
\begin{align}
    \tilde{X}^{anc} = [U_1, U_2]\begin{bmatrix}\Sigma_1& \\ &\Sigma_2\end{bmatrix}\begin{bmatrix}V_1^T\\V_2^T\end{bmatrix}.
\end{align}
Next, setting $Z$ to $U_1$, we construct the matrix $G_i$ such that%
\begin{align}
    G_i = (\tilde{X}^{anc}_i)^{\dagger}U_1.
\end{align}

The described method is summarised in Algorithm \ref{alg:algorithm}

\begin{algorithm}[tb]
\caption{Data Collaboration}
\label{alg:algorithm}
\textbf{Input}\raggedright: $X_i\in{R}^{n_i{\times}m}$, $L_i\in{R}^{n_i{\times}l}$,  individually\\
\textbf{Output}: $\hat{X}\in{R}^{n{\times}l}$, $L\in\mathbb{R}^{n{\times}l}$, $f_i$, $G_i$, $h$, individually\\


\begin{algorithmic}[0] 

\STATE \quad -----USER SIDE----- 
\end{algorithmic}
\begin{algorithmic}[1]
\STATE Generate $X^{anc}\in{R}^{r{\times}m}$ for each user \\ 
\FOR{$i$ in $(1,2,...,d)$}
\STATE Construct $\tilde{X}_i = f_i({X_i})$ and $\tilde{X}^{anc}_i = f_i(X^{anc})$
\ENDFOR
\STATE Centralize $\tilde{X}_i$, $\tilde{X}_i^{anc}$, and $L_i$ for all $i$\\

-----SERVER SIDE-----
\STATE Compute left singular vectors $U_1$ by SVD (6)
\FOR{$i$ in $(1,2,...,d)$}
\STATE Compute $G_i = (\tilde{X}^{anc}_i)^{\dagger}U_1$ (7)
\STATE Compute $\hat{X}_i = \tilde{X}_iG_i$
\ENDFOR
\STATE Set $\hat{X}=[\hat{X}_1^T, \hat{X}_2^T,..., \hat{X}_d^T]^T$, $L = [L_1^T, L_2^T,..., L_d^T]^T$
\STATE Send $\hat{X}$, $L$, $[G_1, G_2,...,G_d]$ back to users\\

-----USER SIDE-----
\STATE Construct model $h$ by a machine learning algorithm useing data $\hat{X}$ as training dataset and $L$ as ground truth so that $h(\hat{X}){\approx}L$

\end{algorithmic}
\end{algorithm}

A notable feature of the Data Collaboration method is that dimensionality reduction function $f_i$ is not shared among the users and all specifications of the function stay secret with each user. As a result, each user obtains a unique machine learning pipeline agnostic of the final classifier, which can be either trained at the server, or at the client side.

\section{Experiments}
\subsection{Experimental Setting}

\begin{figure}
    \centering
    \begin{subfigure}[(a)]{0.475\linewidth}
        \centering
        \includegraphics[width=0.9\linewidth]{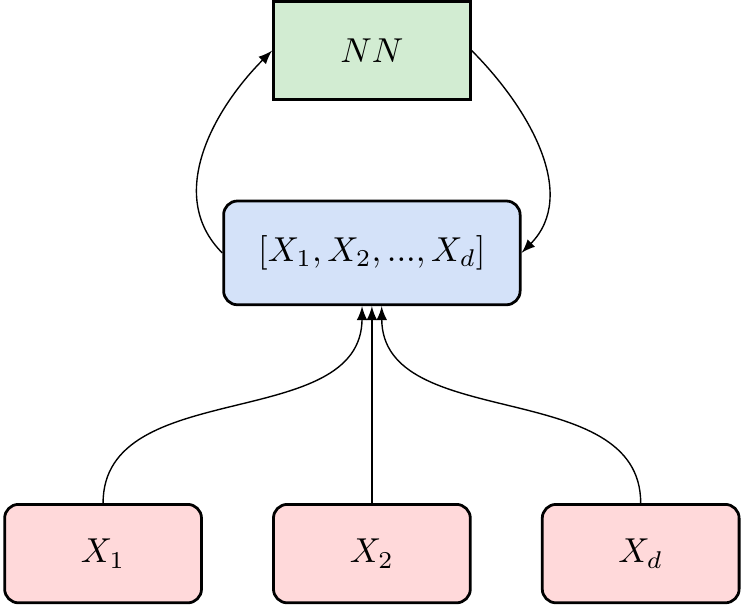}
        \caption[(a)]%
        {Centralized Learning}    
        \label{fig:scheme_centr}
    \end{subfigure}
    \hfill
    \begin{subfigure}[(b)]{0.515\linewidth}  
        \centering 
        \includegraphics[width=0.9\linewidth]{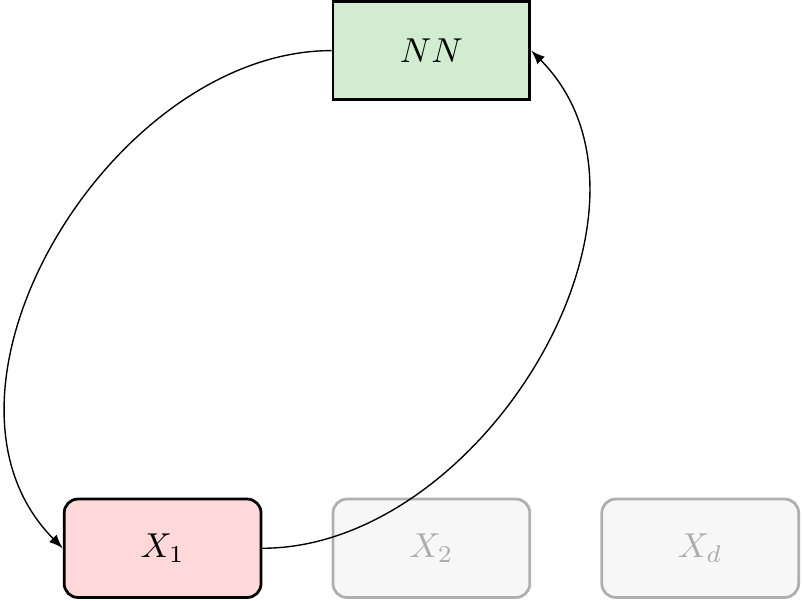}
        \caption[(b)]%
        {Individual Learning}
        \label{fig:scheme_ind}
    \end{subfigure}
    \vspace{8pt}

    \centering
    \begin{subfigure}[b]{0.45\linewidth}   
        \centering 
        \includegraphics[width=0.9\linewidth]{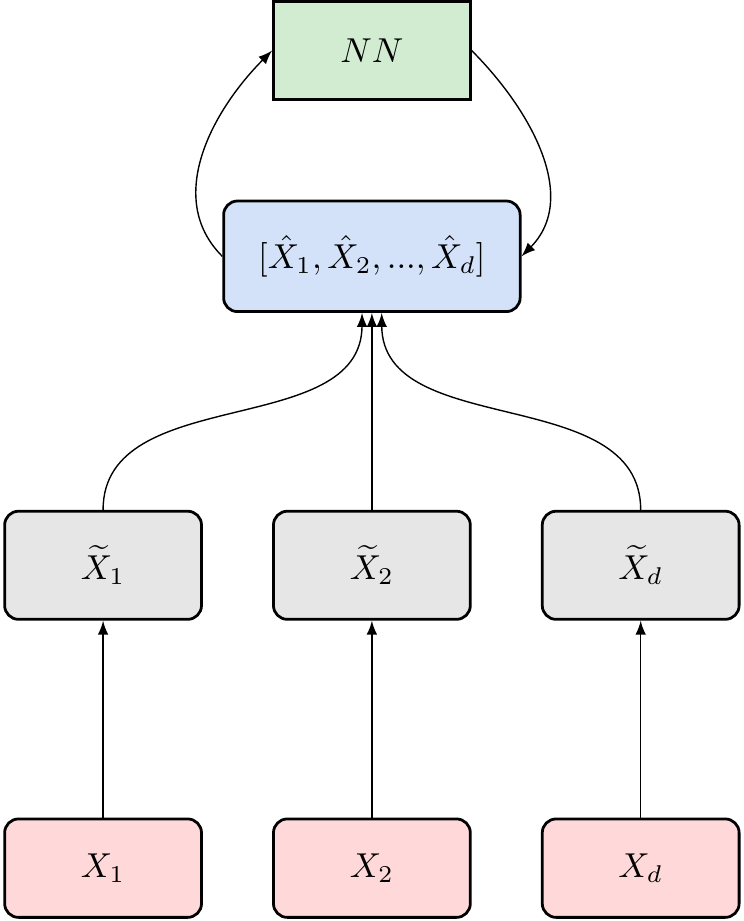}
        \caption[]%
        {Data Collaboration}   
        \label{fig:scheme_dc}
    \end{subfigure}
    \hfill
    \begin{subfigure}[b]{0.535\linewidth}   
        \centering 
        \includegraphics[width=0.9\linewidth]{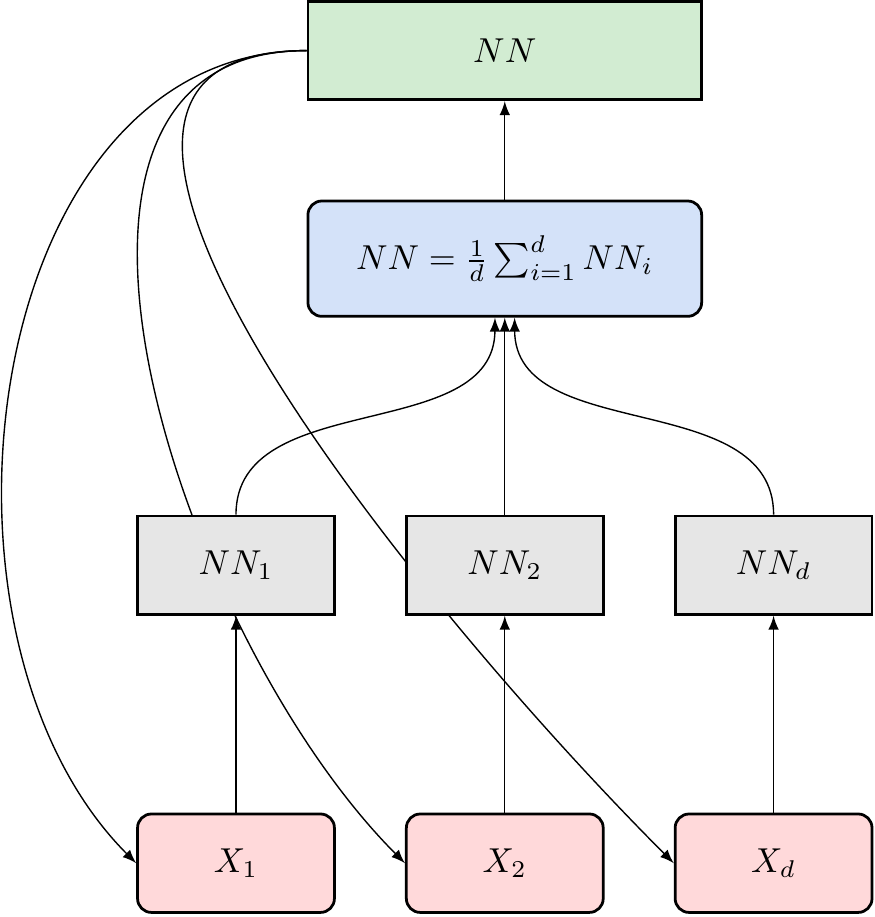}
        \caption[]%
        {Federated Averaging}   
        \label{fig:scheme_fl}
    \end{subfigure}
    \caption[]
    {Schematic representation of the four alternative machine learning workflows compared in this study} 
    \label{fig:scheme_collab}
\end{figure}

In this study, we tested the performance and scalability of Data Collaboration on image classification tasks with two popular datasets, MNIST \cite{lecun2010mnist} and Fashion-MNIST. Both datasets consist of 28x28 pixel grayscale images with 10 class labels. Fashion-MNIST has been recently introduced as a more difficult benchmark dataset than MNIST \cite{xiao2017fashion}. All features were scaled to be between 0 and 1, and images were flattened to 786-dimensional vectors.  

Our implementation of Data Collaboration algorithm differs from the original paper \cite{imakura2020data} in the methods of constructing IR and anchor data. According to the implementation principles of Data Collaboration, IR can be constructed by various dimensionality reduction methods, and the parameters of these methods are not communicated among the users. In practice, for successful collaboration, the choice of the dimensionality reduction method and its parameters should be consistent with the properties of the data. Here, to enable consistent comparison, we unified the IR transformation method across the users.  However, note that even with the same function type $f$, each instance is fitted to the local user's dataset and therefore has different parameters. While the original paper used Kernel-LPP to construct IR's, we chose Singular Value Decomposition (SVD), for its speed and ease of application. The shared anchor data was constructed as a random uniform distribution between 0 and 1. 

We designed two types of collaboration settings, each featuring a different scalability factor: Type 1 for the growing number of collaboration participants (users), and Type 2 of the increasing amount of training data per user. In each instance of the experiment, we recorded validation accuracy on a holdout dataset consisting of 1000 samples, held by the first collaborating user. Unless stated otherwise, the Machine Learning model used for the classification task was a fully-connected neural network with 2 hidden layers of 512 and 128 nodes using ReLu activation function. The training was performed for 24 epochs with batch size 32.

In addition, to test the effectiveness of the collaboration, we compared the results against two baselines: centralized and individual learning. The centralized baseline reflects the best-case scenario when all the data is available for training in one dataset (Figure 1(a)). In contrast, the individual baseline displays the scenario without collaboration when only the data of the first user is used for training (Figure  1(b)). We consider the collaboration successful if the results exceed the individual baseline and approach the centralized baseline in the limit.

All results show the average values over 10 independent runs with random data sampling. The experiments were performed on MacBook Pro with processor of 2.6 GHz Intel Core i7. We used Python 3.7 development environment with Keras deep learning library.

\subsection{Exploration of Data Collaboration parameters}
\begin{figure}
\centering
\includegraphics[width=1\linewidth]{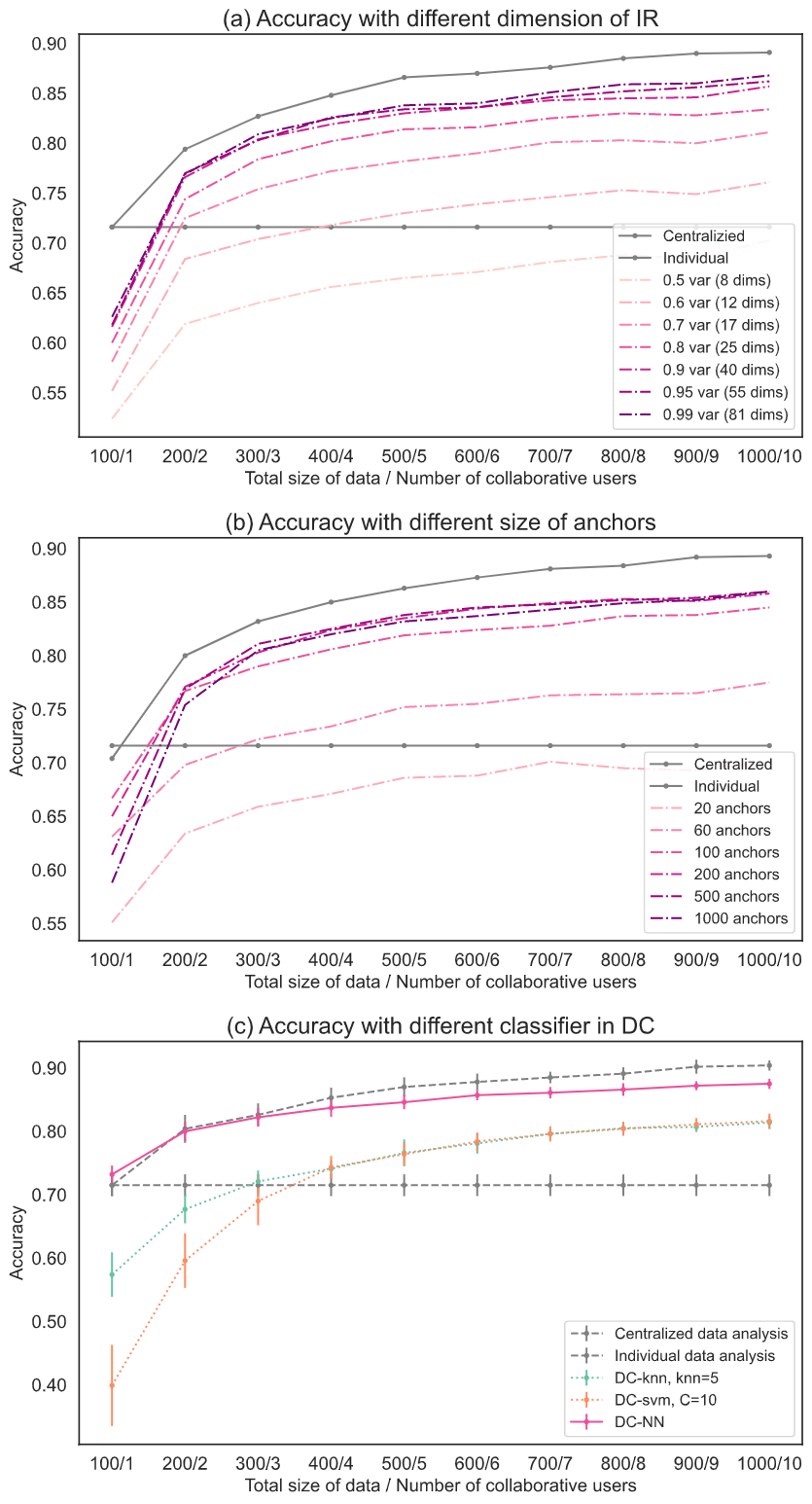}
\caption{Exploration of Data Collaboration Parameters}
\label{fig:params}
\end{figure}

Data Collaboration method has several parameters and conditions which can affect the final results of machine learning tasks. In this experiment, we verify the effects of the two main parameters on the performance of the Data Collaboration algorithm, namely, the dimensionality of the IR created by each user and the amount of anchor data shared by the users. Different parameters were tested on MNIST in Type 1 collaboration setting, first fixing the number of anchors to 500 and varying the dimensions of IR, and then fixing the dimensionality of IR to 50 and changing the number of anchors. In addition, we tested different kinds of classification models fitted on the collaborative data representation: k-Nearest Neighbour Classifier (KNN) with $k=5$ and Support Vector Machine (SVM) with regularization parameter $C=10$ and RBF kernel parametrized by $\gamma=0.01$. The parameter values of the two classifiers were chosen through preliminary cross-validation tests on MNIST data. The results in Figure \ref{fig:params} show that higher dimensionality of IRs yields better results. Similarly, the results of collaboration improve with increasing the amount of shared anchor data. We also observe in Figure 2(c) that although SVM and KNN classifiers achieved worse results than NN, they still outperform individual learning, hence they achieve successful collaboration.

\subsection{Comparison with Federated Averaging in Type 1 collaborative setting}
In the Type 1 collaboration, designed to test how the performance of collaborative learning mechanisms scales with the number of participants, we let each user have 100 samples of data and gradually increased the number of users in collaboration from 1 to 10 users. 

For Data Collaboration (DC), we followed the implementation steps described in Section 2 and schematically presented in Figure 1(c). We set dimensions of IR to 50 and the number of anchors to 500, and Neural Network (NN) as a final classifier, informed by the results of the previous experiment.

Our implementation of Federated Averaging (FedAvg) followed the description of the algorithm in \cite{mcmahan2016communication}. First, the central model is initialized with random weights and passed to each user. Then, users fit the model with their training data for $E$ number of epochs with batch size $B$ and send their model updates back to the server. Consequently, updated from all users are averaged and set as new weights of the central model. This process repeats for $R$ number of rounds. The same structure of the 2-layer fully connected neural network was adopted for the federated model. Furthermore, to match the amount of model training to the centralized baseline, we set $E=1$, $B=32$, and $R=24$. The schematic view of the Federated Averaging algorithm is presented in Figure 1(d).

\begin{figure}
\centering
\includegraphics[width=1\linewidth]{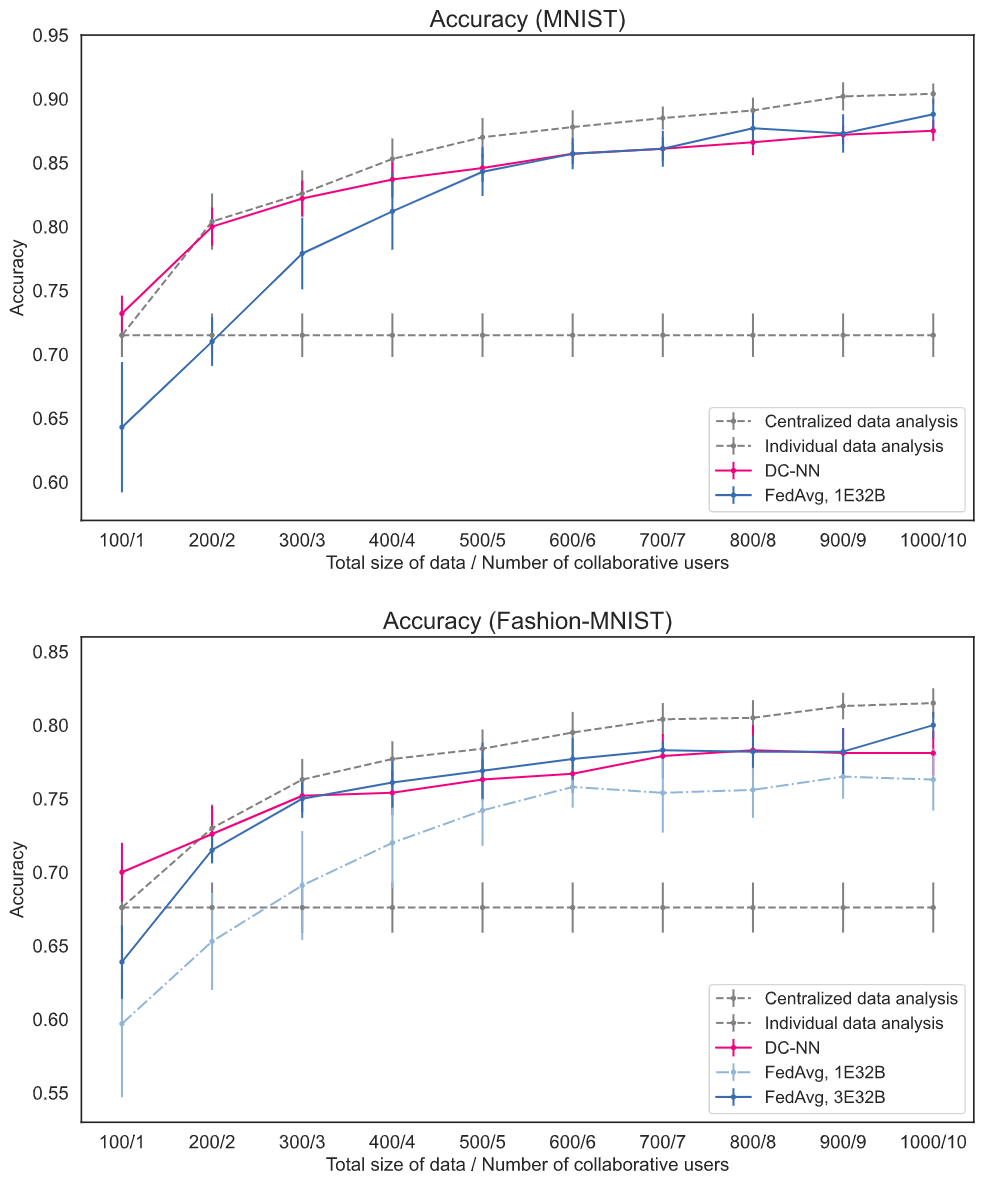}
\caption{Image classification collaboration with increasing number of users}
\label{fig:nuser}
\end{figure}

As shown in Figure \ref{fig:nuser}, on the MNIST dataset, both methods demonstrated similar results approaching the level of centralized machine learning, with FedAvg slightly exceeding the accuracy of DC when the user number grows over seven. On the Fashion-MNIST dataset, DC outperformed FedAvg throughout the user increase stages. Upon observing that FedAvg in configuration $1E32B$ is does not reach convergence on training data, we additionally introduced $3E32B$ configuration of FedAvg, which got compatible results to DC performance. However, it should be noted that total amount of training in FedAvg $3E32B$ now exceeds the training of the centralized baseline. On both datasets, the performance of DC with small user numbers is superior to FedAvg. 

\subsection{Comparison with Federated Averaging in Type 2 collaborative setting}

Having observed in Type 1 collaborative setting that with five users the results of Data Collaboration (DC) and Federated Averaging (FedAvg) become rather close, we then construct a Type 2 collaboration setting, where we fix the number of users to five and gradually increase the amount of data per user from 100 to 1000. For this experiment, we use the same parameters and configurations as in the Type 1 setting, only changing the data sampling. 

\begin{figure}
\centering
\includegraphics[width=1\linewidth]{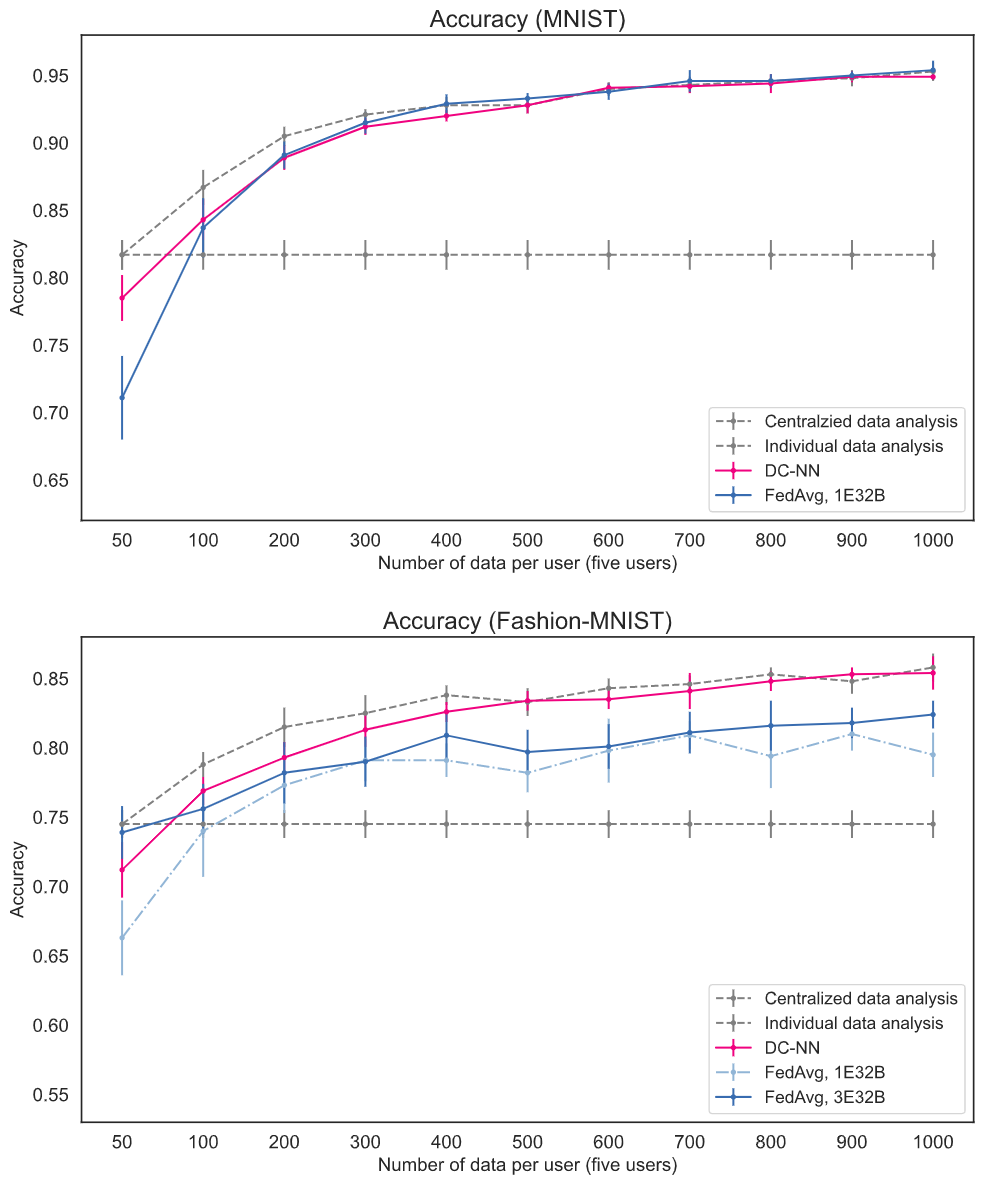}
\caption{Image classification collaboration by five users with increasing the amount of data per user}
\label{fig:ndat}
\end{figure}

The results in Figure \ref{fig:ndat} show that the results of FedAvg and DC on the MNIST dataset continue to be unified, coinciding with centralized learning with the increase of training data. However, on the Fashion-MNIST, FedAvg falls behind DC and displays somewhat unstable behavior in both configurations: $1E32B$ and $3E32B$. It is plausible that with the increased complexity of the dataset and the amount of data per user, FedAvg requires more local epochs per user to reach convergence on training data. 

\section{Discussion}

In the experimental section of this study, we explored different parameters of Data Collaboration method (Figure \ref{fig:params}), and found that higher dimensionality of IRs and larger amount of anchor data yields better results. Because IR acts as feature extraction from the original data, we can expect that the final classification performance can improve when feature extraction is successful but may also degrade if each participant over-reduces dimensionality of their original data. Therefore, we should carefully consider how many features each participant should maintain to guarantee the results. Similarly, anchor data plays a significant role in unifying the IRs, capturing the distance between different projection of the same data points. Hence, anchor data must be sufficiently representative of the original dataset, which can be ensured by increasing the amount of randomly distributed anchor data. The results of KNN and SVM trained on DC representation are interior to DC-NN, although they still perform increasingly better than individual baseline; therefore, they could be used as an alternative to the neural network when the situation requires. 

We also observed that Data Collaboration consistently outperformed Federated Averaging in a small user setting (Figure  \ref{fig:nuser}). This result was expected as Federated Averaging was designed and developed for massively distributed data across over 100,000 users. Moreover, on a more complex image classification task, increasing the amount of data per user required more Federated Averaging training to achieve similar performance to Data Collaboration (Figure \ref{fig:ndat}). 
We attribute the higher performance of Data Collaboration method to the beneficial workings of dimensionality reduction functions underlying the Data Collaboration algorithm, which allows training efficiently with large data samples. Overall, our results show that Data Collaboration is a viable option when model sharing is not desirable.

However, federated learning system without model sharing is a new area of research with a lot of blank spots remaining. For instance, there yet has to be researched and developed suitable IR methods compatible with CNN model training, as well as with different artificial intelligence applications, such as text generation, reinforcement learning, etc. The big open area for research is the privacy of dimensionality reduced representations of data. So far, the Data Collaboration approach only guarantees that original data cannot be reconstructed or approximated from Intermediate Representations shared by users, hence, it protects the original data leakage from semi-trusted participants. Still, further research is needed to explore statistical privacy guarantees. Having said that, there are works in literature that use dimensionality reduced representations of data for privacy-preserving applications \cite{liu2020privacy}, \cite{alotaibi2012non}, \cite{nguyen2020autogan}.

From the implementational point of view, we identify two types of scenarios for collaborative machine learning, Service Provider Collaboration and Competitor Collaboration, and consider the applicability of model sharing versus non-model sharing collaboration in each of them. In the first scenario, most common for Federated Learning applications, there is a service provider who wants to use users' data to improve the service application. In this case, model sharing is entirely suitable as the provider wants to guarantee the same model quality to all users, and the communication infrastructure necessary to train the model can be easily implemented within the service. On the other hand, deploying Data Collaboration in this scenario might be challenging because it would require producing IRs independently by each user. 

In the second scenario, several competing companies might decide to form data partnerships to develop better prediction models for their businesses. Examples of such partnerships are common in banking, insurance, security, and medical sectors, were all parties can benefit from obtaining external data for training predictive models, yet the parties are competing in providing the services. For instance, a retail company may want to obtain insights of customer experience from other retails, yet keep their sales prediction model private. In this case, model sharing scheme can disincentivize collaboration, while a Data Collaboration scheme would allow each party to add value to model development over the competitors. Data Collaboration approach, in this case, can offer more balanced incentives to all parties and would be easier to implement as it happens in "one pass," as opposed to iterative communications required by conventional federated learning systems. 

\section{Conclusion}
In this paper, we explored an alternative federated learning system called Data Collaboration. The distinct characteristic of this method is that the users do not share the final result of the collaboration in the form of a predictive machine learning model. Instead, they independently create a dimensionality reduced representation of their data and use shareable anchor data to integrate these representations into one dataset. The product of this federated learning system is a combined dataset in a lower-dimensional space and unique transformation functions distributed to each user that can be applied to new data samples for training machine learning models. To test the effectiveness of Data Collaboration, we compared its performance to three alternative frameworks: centralized machine learning, individual machine learning, and federated learning, represented by Federated Averaging method.

We showed that Data Collaboration can achieve comparable results to Federated Averaging, as well as to a centralized training setting, which makes it a practical alternative to conventional methods. Moreover, in our experiments, Data Collaboration consistently outperformed Federated Averaging in a small user setting and required less training to achieve same results on bigger datasets.

We conclude that Data Collaboration approach can be a viable option in scenarios where conventional federated learning with model sharing is problematic. Such scenarios include but not limited to competitor collaborations, where each user is motivated to produce better results then the rest. Moreover, the dimensionality reduction step of the Data Collaboration algorithm exposes interesting properties of improving machine learning performance while protecting privacy through perturbing original data and thus deserves closer attention of federated learning community. 

Further directions of research into the Data Collaboration approach include the formal privacy analysis of dimensionality reduced representations of data, as well as the expansion of experimental baselines to include larget number of users, more complex model architectures and non-iid data distribution scenarios. 

\section*{Acknowledgments}
The present study is supported in part by the Japan Science and Technology Agency (JST), ACT-I (No. JPMJPR16U6), the New Energy and Industrial Technology Development Organization (NEDO) and the Japan Society for the Promotion of Science (JSPS), Grants-in-Aid for Scientific Research (Nos. 17K12690, 18H03250, 19KK0255).

\bibliographystyle{named}
\bibliography{ijcai20}

\end{document}